\pdfoutput=1
\documentclass[10pt,twocolumn,letterpaper]{article}

\usepackage{cvpr}
\usepackage{times}
\usepackage{epsfig}
\usepackage{graphicx}
\usepackage{amsmath}
\usepackage{amssymb}

\usepackage{bm}
\usepackage{float}
\usepackage[utf8]{inputenc}
\usepackage[english]{babel}
\usepackage{amsthm}
\usepackage{subfigure}
\usepackage{algorithmic}
\usepackage{multirow}
\usepackage{makecell}
\usepackage{stfloats}

\usepackage[lined,boxed,commentsnumbered,ruled]{algorithm2e}

\cvprfinalcopy
\def\cvprPaperID{}

\begin{document}

\title{3D Objectness Estimation via Bottom-up Regret Grouping}

\author{Zelin Ye$^{1}$, Yan Hao$^{1}$, Liang Xu$^{1}$, Rui Zhu$^{1}$, Cewu Lu$^{1}$\\
	$^{1}$ Shanghai Jiao Tong University \\
{\tt\small h\_e\_r\_o@sjtu.edu.cn}, {\tt\small honeyhaoyan@sjtu.edu.cn}, \\ {\tt\small liangxu@sjtu.edu.cn}, {\tt\small zirconium@sjtu.edu.cn}, {\tt\small lucewu@sjtu.edu.cn}
}

\maketitle
\begin{abstract}
   3D objectness estimation, namely discovering semantic objects from 3D scene, is a challenging and significant task in 3D understanding. In this paper, we propose a 3D objectness method working in a bottom-up manner. Beginning with over-segmented 3D segments, we iteratively group them into object proposals by learning an ingenious grouping predictor to determine whether two 3D segments can be grouped or not. To enhance robustness, a novel regret mechanism is presented to withdraw incorrect grouping operations. Hence the irreparable consequences brought by mistaken grouping in prior bottom-up works can be greatly reduced. Our experiments show that our method outperforms state-of-the-art 3D objectness methods with a small number of proposals in two difficult datasets, GMU-kitchen and CTD. Further ablation study also demonstrates the effectiveness of our grouping predictor and regret mechanism.
\end{abstract}

\section{Introduction}

Recently, 3D understanding \cite{3d-scene-1, squeezeseg, voxelnet} has achieved remarkable progress and greatly advanced the intelligence of other domains, \eg robotic manipulation \cite{robot-1} and automatic drive \cite{pixor}. Many related topics such as 3D recognition, 3D object detection have been extensively studied by researchers. Beyond those topics, 3D objectness is relatively less explored. It aims at discovering semantic objects of 3D scenes without knowing object categories in advance. This can serve as a more general tool under many circumstances, \eg a robot working in a unknown and open environment. Therefore, this work seeks to in-depth study this problem and propose a desirable solution.

Classical 3D objectness works mainly consist of two branches. On the one hand, with the high variety of effective 2D objectness algorithms or 2D region proposal network architectures, some methods \cite{avod} intuitively conduct an extension of 2D methods to 3D space. However, the huge 3D search space and the irregular format of point clouds make such simple extension inefficient and difficult to achieve expected results. Besides, such 2D-based methods \cite{f-pointnet} might omit some objects that could only be clearly detected from 3D space.

On the other hand, some methods \cite{3d-ss} directly conduct objectness from 3D perspective. By making full use of the voxel structure, they develop various similarity functions, thus perform bottom-up grouping. Nevertheless, such hand-crafted metrics require large quantities of labor and trials. Besides, the scalability of such methods is also limited. Recent 3D objectness methods \cite{gmu} generate proposals from dense point clouds via multi-view approaches and achieve promising performance. However, in practice, it is often difficult and laborious to obtain the multi-view perceive or even the dense point clouds of the whole scene.

\begin{figure}[t]
	\hspace{0.03\linewidth}
	\subfigure[RGB image]{
		\includegraphics[width=0.4\linewidth]{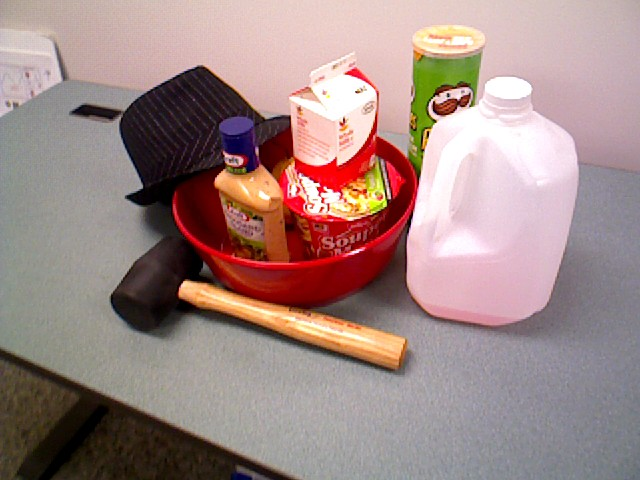}
		\label{fig:intro-a}
	}
	\hspace{0.05\linewidth}
	\subfigure[3D segments]{
		\includegraphics[width=0.4\linewidth]{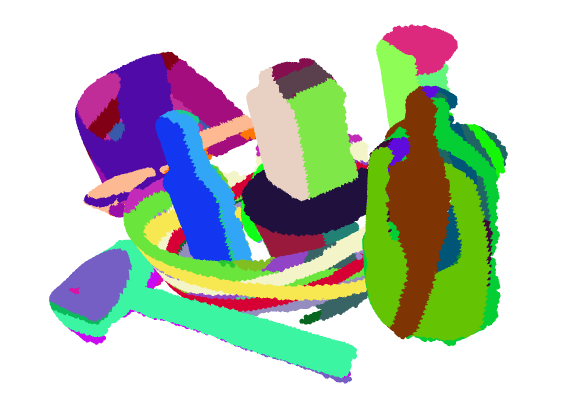}
		\label{fig:intro-b}
	}

	\hspace{0.03\linewidth}
	\subfigure[Proposed Candidates]{
		\includegraphics[width=0.4\linewidth]{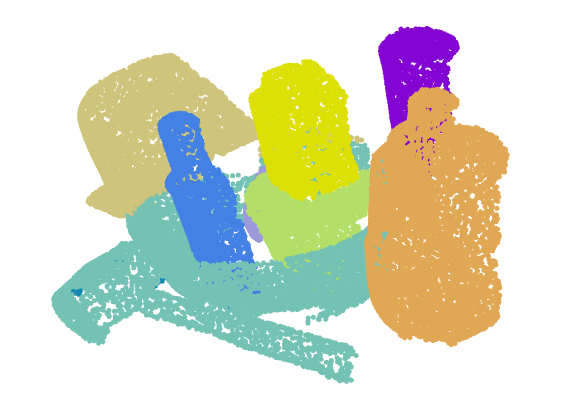}
		\label{fig:intro-c}
	}
	\hspace{0.05\linewidth}
	\subfigure[Ground-truth]{
		\includegraphics[width=0.4\linewidth]{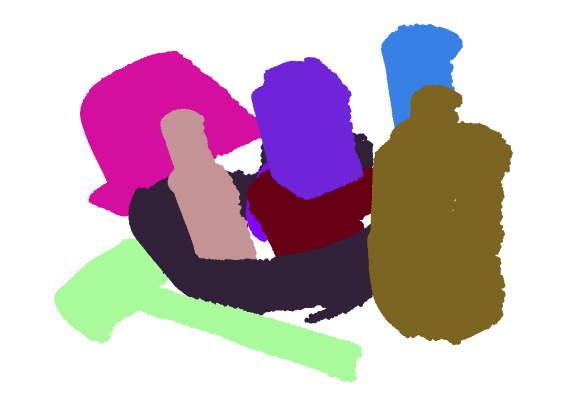}
		\label{fig:intro-d}
	}
	
	\caption{A typical result of our objectness method on CTD \cite{ctd}. Fig. \ref{fig:intro-a} is the RGB image of the scene and \ref{fig:intro-b} shows corresponding 3D segments produced by over-segmentation. \ref{fig:intro-c} and \ref{fig:intro-d} are detected 3D proposals by our method and the ground-truth respectively.}
	\label{fig:intro}
\end{figure} 

In this work, we present a bottom-up objectness method on 3D point clouds. We first decompose the scene into a set of over-segmented 3D segments, then iteratively group them. In this way, a grouping predictor using neural network is introduced to estimate the probability that a pair of 3D segments belong to a same object. It learns from a large amount of 3D object models to extract the spatial features of each pair of 3D segments and make the final estimation.

We also notice that prior bottom-up grouping algorithms \cite{selective-search} work in a greedy manner and lack a global view. Therefore, we propose a novel regret grouping mechanism. By considering larger semantic space, namely relationship among all adjacent 3D segments beyond the target pair, the occurrence of mistaken grouping can be effectively reduced. The final proposals are achieved when no 3D segment pair is predicted as the same object.

We evaluate our objectness method in two datasets, GMU-kitchen dataset \cite{gmu} and Cluttered Table-top Dataset (CTD) \cite{ctd}. The experimental results show that our method outperforms state-of-the-art 3D proposal generation methods. In particular, we also demonstrate its robustness with the assistance of our regret grouping mechanism, especially in highly occluded scenes.

The main contributions of this work are as follows:

\begin{enumerate}
	\item We present an effective 3D objectness method, which utilizes only spatial information of 3D point clouds and avoids hand-crafted metrics (\eg various similarity functions).
	
	\item We introduce regret grouping mechanism to largely address the mistaken grouping issue due to the greedy scheme. Some mistakenly grouped 3D segments can be regretted via more comprehensive analysis.
	
	\item We show that our objectness method outperforms state-of-the-art 3D objectness methods in two difficult datasets of 3D scenes.
\end{enumerate}

\section{Related Work}

\noindent{\bf Object Detection.} Object detection \cite{object-detection-review} on 2D images has enjoy great popularity \cite{YOLO, ssd, faster-rcnn} and shown its strength in many fields, such as image classification \cite{krizhevsky2012imagenet}, face recognition \cite{yang2016multi} and human behavior analysis \cite{human_activity}. Based on a rich literature of existing 2D object detection techniques, many works \cite{voxelnet, avod} try to adopt two-stage detector in 3D space, namely generating object proposals and then classifying them into different categories. However, the enormous number of proposals generated by conventional schemes (\eg sliding windows) prevents them from real-time applications. An effective objectness method is hence in urgent demand.

In this work, we focus on generating reasonable object proposals in 3D space, which can be further applied to object detection and then serves in more real-world tasks as a general tool.

\noindent{\bf 2D Objectness.} Objectness is initially conducted in 2D image space, aiming at generating object candidates from visual clues such as texture and contour features. There are three branches of approaches to executing 2D objectness \cite{object-detection-review}. The first kind of methods are based on region proposals selection \cite{rcnn, SPP-net, fast-rcnn}. This kind of models have made much progress recently, however, their speed is restricted by the number of region proposals. The second kind of methods take images as a whole and treat 2D objectness estimation as a regression problem \cite{attentionnet, YOLO, ssd, YOLOv2}, which is usually limited to specific kinds of objects set in advance. The third kind of methods make use of salience, that is, the noticeable visual information in an image, including FT \cite{FT}, GC \cite{GC}, SF \cite{SF}, and PCAS \cite{FCAS}, which may be inefficient and not robust when it comes to complicated circumstances.

\noindent{\bf 3D Objectness.} Recently, the use of spatial information in 3D space \cite{pointnet, pointnet++} has been very popular and constantly applied to various domains, such as robotic manipulation. Therefore, an efficient method of creating object candidates in 3D space is urgently required. However, due to the irregular data formats of 3D point clouds, 3D objectness is not yet well resolved.

There are two main kinds of methods to deal with the problem. One is to project 3D data to 2D view images and then apply 2D methods to extract features \cite{chen2017multi, li2016vehicle}. The other is to directly extract 3D features and then convert 3D feature map into 2D space \cite{li20173d, prokhorov2010convolutional, yan2018second, voxelnet}. However, both kinds of methods are actually dealing with the problem in conventional ways and cannot fully utilize unique advantages in 3D representation. Some methods \cite{voxelnet, sgpn} also aggregate objectness in end-to-end framework of some specific vision tasks (\eg object detection, instance segmentation), known as RPN. Nevertheless, the training of RPN is highly affected by the whole architecture and it is hard to separate it out and train it alone.

There are also some works that conduct over-segmentation in 2D space and then group the segmented parts in 3D space greedily \cite{3d-ss, 2d-segmentation-3d-merge}. However, quantities of hand-crafted metrics are highly demanded and the generalization ability is also limited in such methods. In this work, we introduce a manner that directly conducts over-segmentation in 3D space and group the segmented parts refer to spatial features extracted from neural networks instead of similarity functions. Furthermore, incorporate with our regret mechanism, we can withdraw most mistaken grouping operations caused by classical greedy manner, thus reducing incorrect proposals.

\section{Method}

\begin{figure*}[!t]
	\centering
	\includegraphics[width=0.95\linewidth]{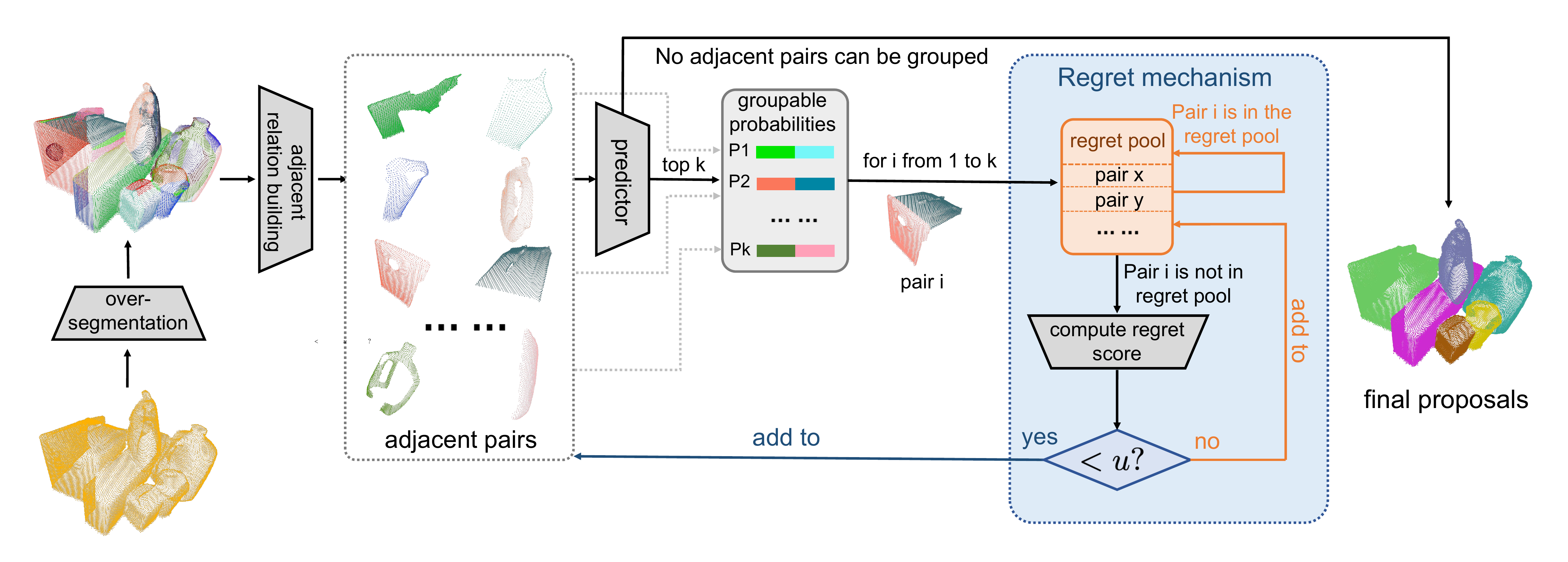}
	\caption{Overview of our 3D objectness framework. We first over-segment the point cloud of a given scene and thus extract adjacent pairs of 3D segments. Then we group the top $k$ pairs with the largest groupable probabilities estimated by a grouping predictor. In particular, potential mistaken grouping operations can be backtracked via a regret mechanism, which is elaborated in Sec. \ref{sec:regret}.}
	\label{fig:pipeline}
\end{figure*}

\subsection{Overview}
\label{sec:overview}
Our objectness method is a bottom-up grouping process. Firstly, we conduct an over-segmentation on the 3D point cloud of a given scene and obtain a set of 3D segments (see Sec. \ref{sec:overseg}), then adjacent 3D segments are iteratively grouped into larger ones by making use of a learned grouping predictor (see Sec. \ref{sec:predictor}) until no 3D segments can be grouped. Besides, to enhance robustness, we also introduce a regret mechanism to withdraw incorrect grouping operations (see Sec. \ref{sec:regret}). The overall pipeline is illustrated in Fig. \ref{fig:pipeline}.

\subsection{Over-segmentation}
\label{sec:overseg}
\begin{figure}[!htbp]
	\centering
	\includegraphics[width=0.99\linewidth]{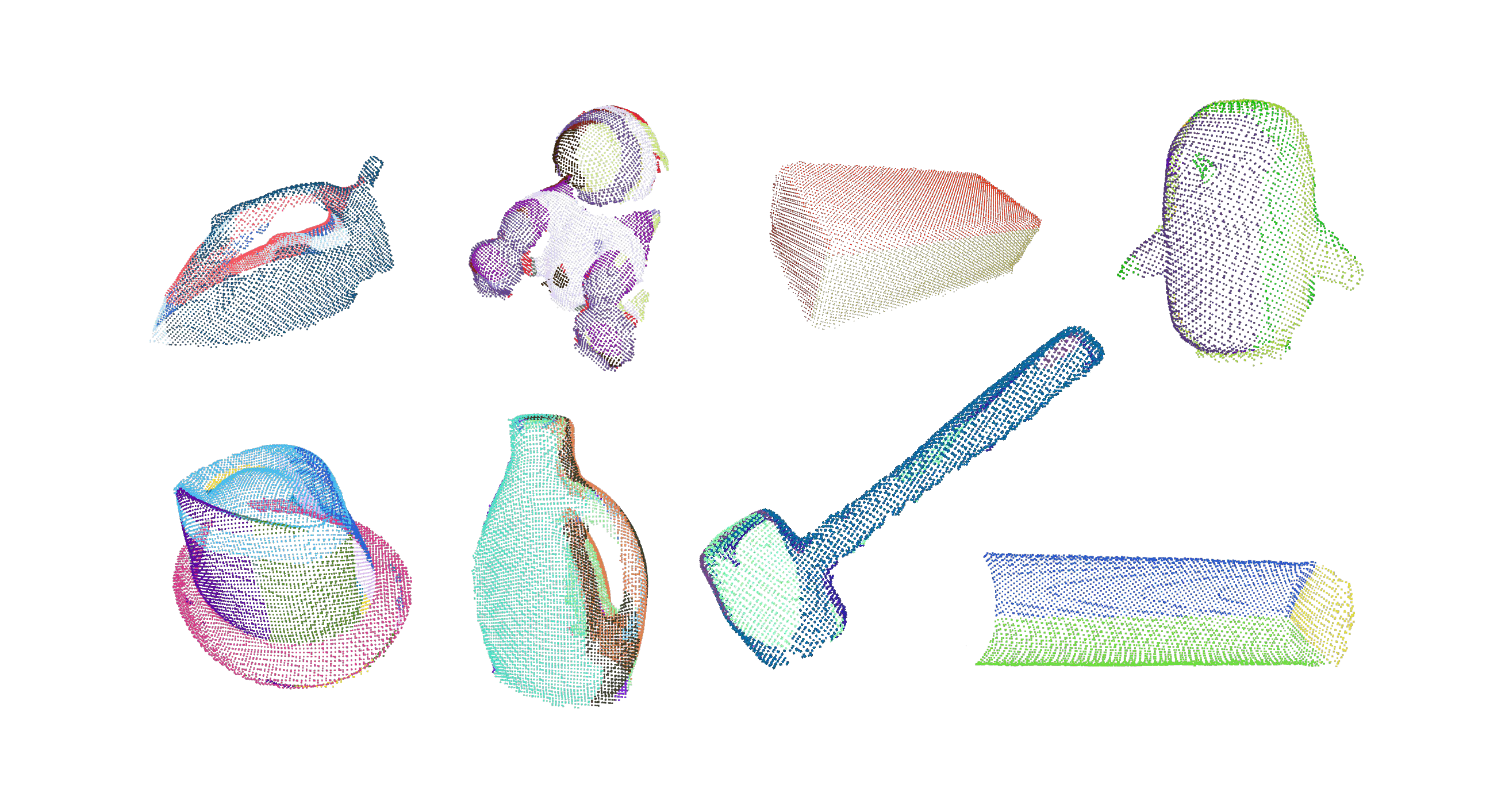}
	\caption{The over-segmentation results on some complicated 3D point cloud structures.}
	\label{fig:segment-plane}
\end{figure}

Since our pipeline is a bottom-up process, we first over-segment the input 3D point clouds into a set of 3D segments for later grouping procedure. In general, any 3D point cloud structures can be segmented into planes (Fig. \ref{fig:segment-plane}). In this over-segmentation procedure, we continuously apply RANSAC \cite{ransac} algorithm to search planes and take them as the initial 3D segments for subsequent grouping process.

\begin{figure}[htbp]
	\centering
	\includegraphics[width=0.99\linewidth]{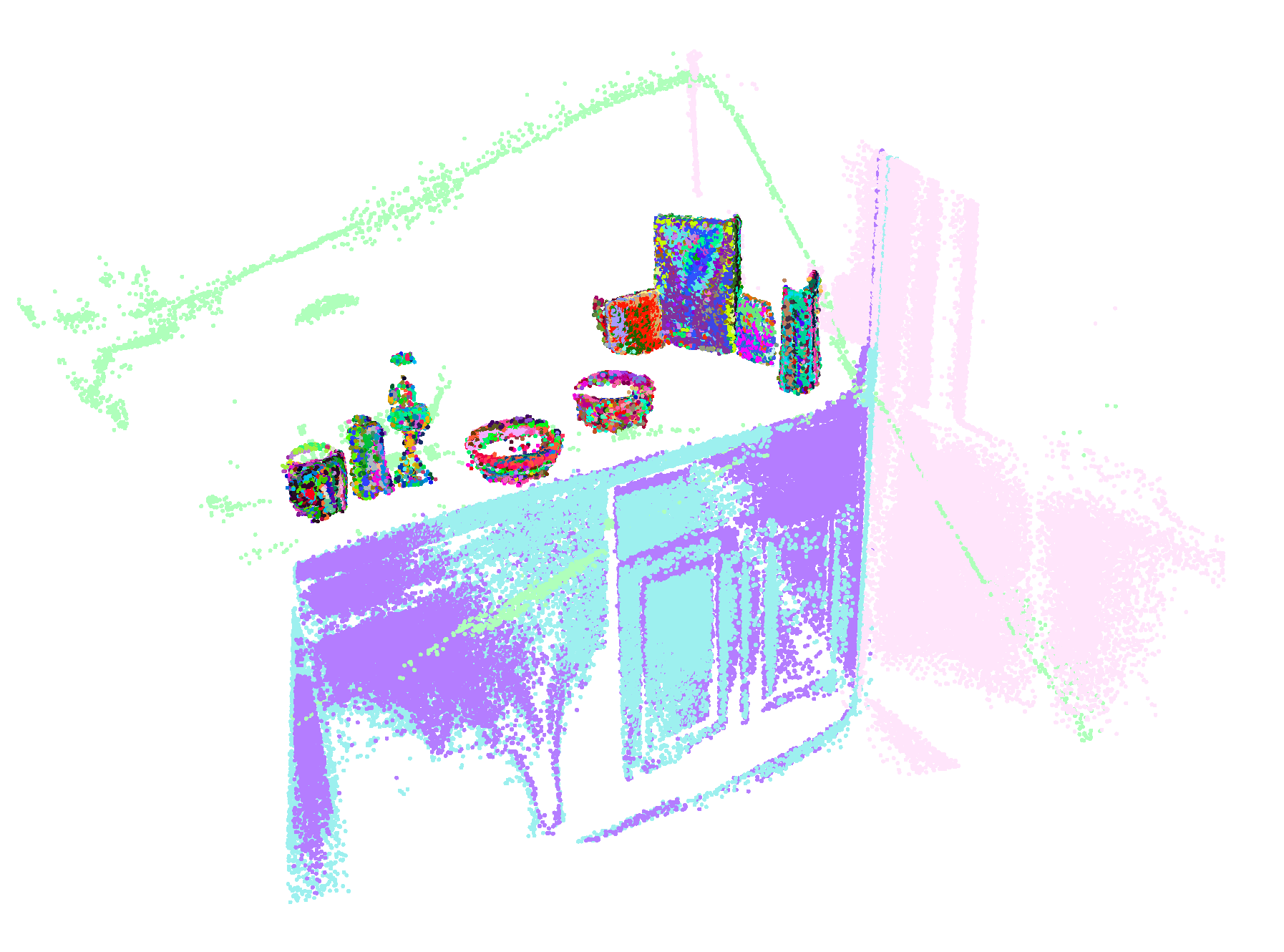}
	\caption{Over-segmentation results of a typical scene of GMU-kitchen dataset \cite{gmu}.}
	\label{fig:over-seg}
\end{figure}

This initial over-segmentation can largely ensure the points in the same 3D segment belong to the same semantic object. Our experiment on CTD dataset shows that the above assumption holds for $99.7\%$ segments. Note that we remove the background of large planes (\eg desktop, wall) to improve efficiency and avoid mistakes. A typical over-segmentation result is illustrated in Fig. \ref{fig:over-seg}.

\subsection{Adjacent Relationship Building}
Basically, the grouping operations are only executed for adjacent 3D segments. A simple yet effective adjacent relationship building scheme is proposed: we partition the whole point cloud into a $m \times m \times m$ voxel-grid ($m$ is $256$ in our experiments). If two 3D segments have points in the same grid, we label them as an adjacent 3D segment pair. In the grouping process, we will skip pairs of 3D segments that are not adjacent, which can accelerate the training process and enhance robustness.

\subsection{Learn to Group}
\label{sec:predictor}

\begin{figure}[t]
	\centering
	\includegraphics[width=0.99\linewidth]{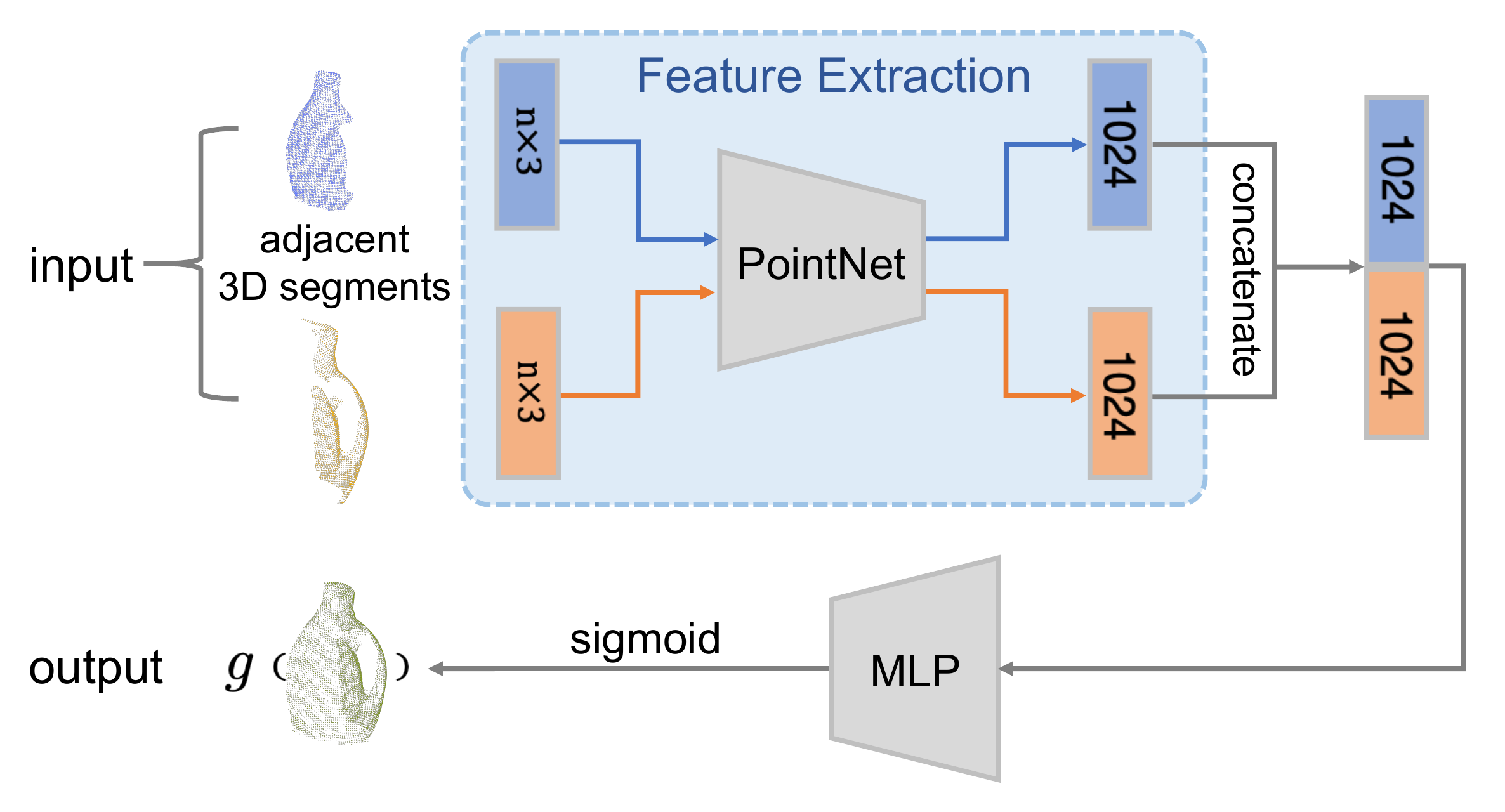}
	\caption{The architecture of our grouping predictor. The point clouds of a pair of adjacent 3D segments are fed to PointNet to extract features. An MLP serves as a prediction module to estimate the groupable probability. Batch normalization is adopted in both convolutional and fully-connected layers.}
	\label{fig:architecture}
\end{figure}

Given initial 3D segments obtained by over-segmentation process, we iteratively group them into object proposals. In this way, a criterion is required to determine whether two adjacent 3D segments belong to the same object or not. Instead of adopting conventional grouping schemes that rely on hand-crafted grouping metrics, we seek to construct a grouping predictor and learn such metrics from a large amount of 3D semantic object models. The predictor consists of two modules as illustrated in Fig. \ref{fig:architecture}. We first apply a feature extraction module to obtain spatial features of two 3D segments, then a prediction module is followed to estimate the corresponding groupable probability.

\paragraph{Feature extraction module.}
We first apply uniform sampling and zero padding to all 3D segments to ensure the input dimension as $n \times 3$, where $n$ is 2048 in our experiments and $3$ denotes the 3D Cartesian coordinates. PointNet \cite{pointnet} is employed as the backbone to extract the spatial feature of each 3D segment. Note that we found that PointNet feature is effective enough to handle this problem, since most of 3D segments are relatively simple, and there is no need to involve more complex models such as PointNet++ \cite{pointnet++}.

\paragraph{Prediction module.}
We get two aggregated feature vectors from the previous module. They are concatenated and then fed to a 5-layer MLP with the number of neurons (2048, 1024, 512, 128, 32) to estimate the final probability that two 3D segments belong to the same object. This module is ended with a \textit{Sigmoid} function to compress the output into the range of 0 to 1.

The training of our grouping predictor is conducted on large-scale 3D object datasets \cite{ctd, gmu}. We produce 3D segments on those datasets and label two adjacent 3D segments on the same object as 1, otherwise as 0. Given a 3D pair $p_i$ of segments $\mathcal{A}$ and $\mathcal{B}$, the corresponding prediction is denoted as $g(p_i)$ or $g(\mathcal{A}, \mathcal{B})$.

\paragraph{Loss function.}
Intuitively, the problem can be considered as a regression problem and Huber loss \cite{huber-loss} can be adopted. However, since the ground-truth is binary, the loss functions for regression tend to make the predictions infinitely close to 0 or 1, especially in the case of uneven distribution of training data. We address this issue with binary cross-entropy loss (Eqn. \ref{eqn:loss}) and $\ell^2$ regularization is adopted.
\begin{equation}
L_i = -w_i[y_i \log g(p_i) + (1 - y_i) \log (1 - g(p_i))],
\centering
\label{eqn:loss}
\end{equation}
where $y_i$ is the ground-truth of $i$-th pair of 3D segments $p_i$ and $w_i$ is the weight of $i$-th item, which is identical in our experiments.

\subsection{Regret Grouping}
\label{sec:regret}
As described in Sec. \ref{sec:overview}, our objectness method is treated as a bottom-up grouping process. The standard bottom-up method is selective search \cite{selective-search} that works in a greedy manner. Conventional selective search just discards grouped segments after each iteration, which is intuitive and easy to operate. Nevertheless, once an incorrect grouping is made, the subsequent grouping process would suffer from increasing biases, resulting in many incorrect proposals. To enhance the fault tolerance of our method, a \textit{\textbf{regret mechanism}} is presented. For each grouped 3D segment pair, we compute a regret score to determine whether they should be regretted or not. In what follows, we introduce the definition of regret score, thus present our regret mechanism.

\paragraph{Regret score.}
Suppose two 3D segments $\mathcal{A}, \mathcal{B}$ have been grouped as a new 3D segment $\mathcal{D}$, then we verify the correctness of this grouping by considering the relationship between $\mathcal{D}$ and other 3D segments adjacent to it. By assuming there is a 3D segment $\mathcal{C}$ adjacent to $\mathcal{B}$, three events are defined as the following:

\begin{itemize}
	\item $E_1$: $\mathcal{C}$ and $\mathcal{B}$ belong to the same object.
	\item $E_2$: $\mathcal{C}$ and $\mathcal{D}$ belong to the same object.
	\item $E_3$: $\mathcal{A}$ and $\mathcal{B}$ do not belong to the same object (we should regret the grouping of $\mathcal{A}$ and $\mathcal{B}$ in this case).
\end{itemize}

In the viewpoint of probability, if $P(E_1) = 1$ and $P(E_2) = 0$, we can infer $P(E_3) = 1$. Proof by contradiction is adopted.

\begin{proof}
	Suppose $P(E_3) \neq 1$, that is $\mathcal{A}$ and $\mathcal{B}$ possibly belong to the same object. Because $P(E_1) = 1$, we have $\mathcal{A}$, $\mathcal{B}$ and $\mathcal{C}$ possibly belong to the same object. $\mathcal{D}$ is the union of $\mathcal{A}$ and $\mathcal{B}$, therefore, $\mathcal{C}$ and $\mathcal{D}$ possibly belong to the same object. That is $P(E_2) \neq 0$. This contradicts the fact that  $P(E_2) = 0$. Hence $P(E_3) = 1$.
\end{proof}

\begin{figure}[htbp]
	\centering
	\subfigure[Scene layout]{
		\includegraphics[width=0.53\linewidth]{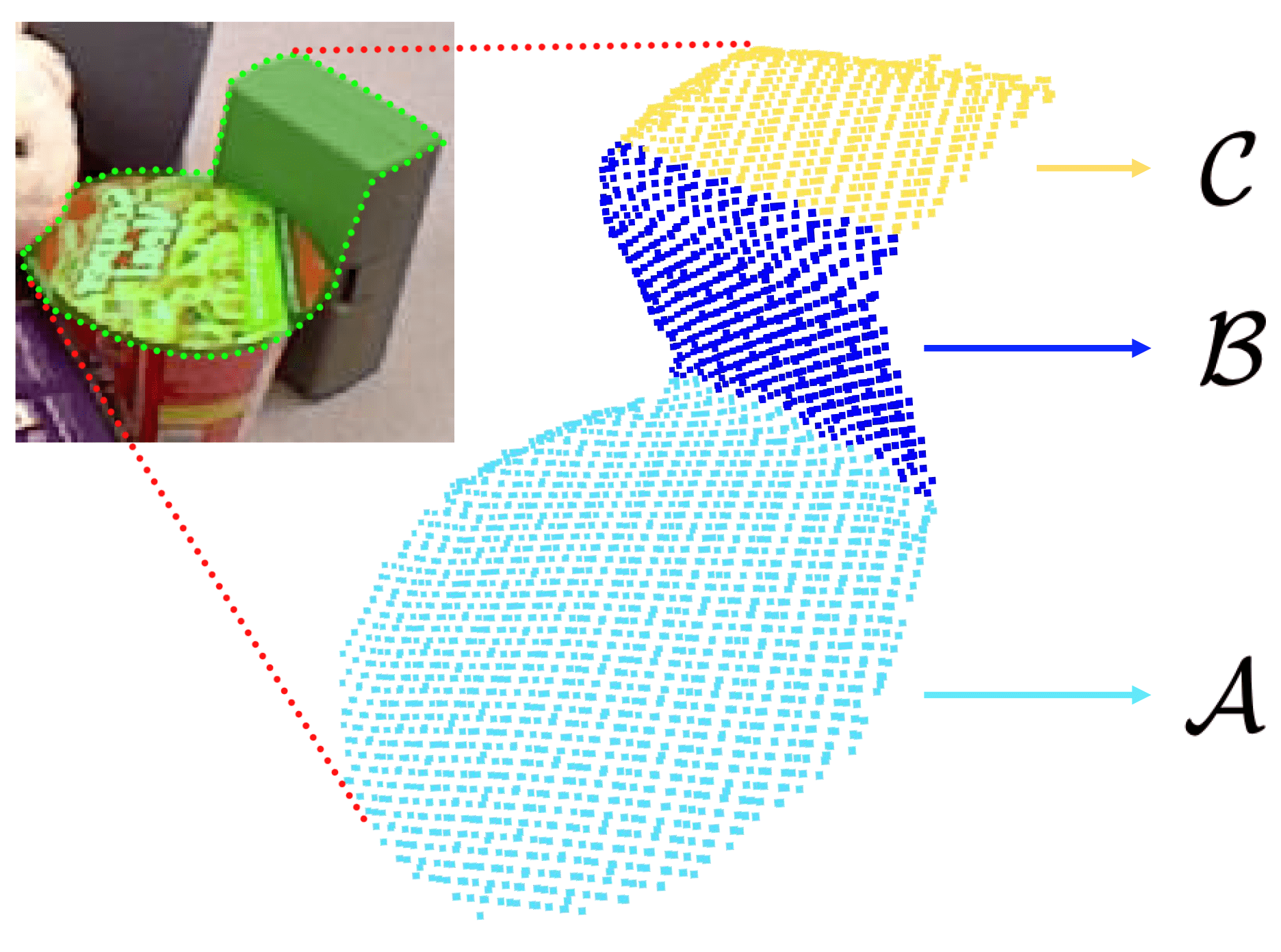}
		\label{fig:regret_sample1_scene}
	}
	\subfigure[Prediction results]{
		\includegraphics[width=0.41\linewidth]{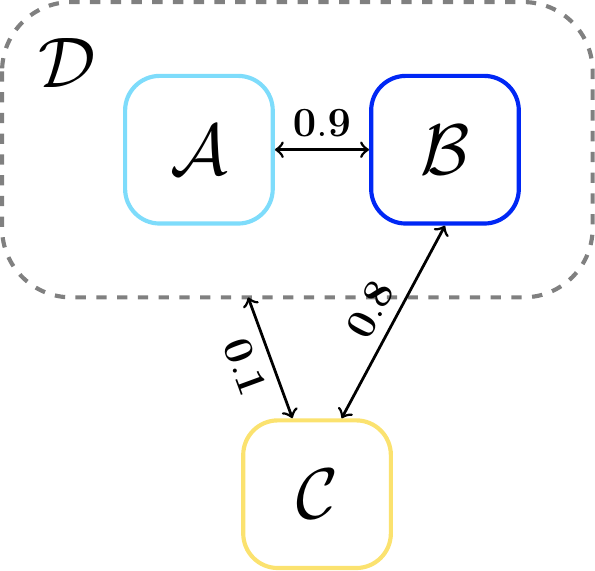}
		\label{fig:regret_sample1_anno}
	}
	\caption{A specific case where regret is needed. Fig. \ref{fig:regret_sample1_scene} shows that $\mathcal{B}$ and $\mathcal{C}$ belong to the same object while $\mathcal{A}$ and $\mathcal{B}$ not. The numbers in Fig. \ref{fig:regret_sample1_anno} denote the prediction results. Since $g(\mathcal{C},\mathcal{B}) - g(\mathcal{C},\mathcal{D})=0.7$, which is a relatively large value, we tend to regret the grouping of $\mathcal{A}$ and $\mathcal{B}$.}
	\label{fig:regret_sample1}
\end{figure}

Therefore, we regret the grouping of $\mathcal{A}$ and $\mathcal{B}$, when $P(E_1)$ is large, and in the meanwhile $P(E_2)$ is small. In practice, $P(E_1)$ and $P(E_2)$ can be intuitively indicated by $g(\mathcal{C},\mathcal{B})$ and $g(\mathcal{C},\mathcal{D})$ respectively. A large $g(\mathcal{C},\mathcal{B}) - g(\mathcal{C},\mathcal{D})$ implies that $P(E_2)$ is approximate to 1, so we need to regret the grouping of $\mathcal{A}$ and $\mathcal{B}$ as illustrated in Fig. \ref{fig:regret_sample1}.

Furthermore, we consider the case of multiple neighborhoods, where the regret score can be written as
\begin{equation}
\begin{gathered}
s = \max \{s_1, s_2\}\\
s_1 = \max_{\mathcal{V} \in \Theta(\mathcal{B})} g(\mathcal{V},\mathcal{B}) - g(\mathcal{V},\mathcal{D}).\\
s_2 = \max_{\mathcal{V} \in \Theta(\mathcal{A})} g(\mathcal{V},\mathcal{A}) - g(\mathcal{V},\mathcal{D}).
\end{gathered}
\label{eq:regret_score}
\end{equation}
where $\Theta(\mathcal{A})$ and $\Theta(\mathcal{B})$ are the neighborhood sets of $\mathcal{A}$ and $\mathcal{B}$, respectively.

\paragraph{Regret mechanism.}
	With the regret score metric, we present our regret mechanism. The initial 3D segments are produced by over-segmentation. In each iteration, we find the adjacent pairs with the highest $k$ groupable probabilities and form them as a set of grouping candidates.
	
	For each candidate, if its regret score is larger than a certain threshold, we regret this grouping and put the corresponding pair into a regret pool. Otherwise, they would be really grouped and discarded in later iterations. Note that we directly regret grouping for pairs already staying in regret pool without computing their regret scores, thus avoiding repeated computation and promote efficiency.
	
	When a 3D segment can not find any other 3D segments to group with, namely all the groupable probabilities are less than a threshold, we output it as an object proposal. The overall regret grouping algorithm is sketched in Alg\onedot \ref{alg:grouping-t}.
	
	\begin{algorithm}[htbp]
		\SetKwInOut{Input}{Input}
		\SetKwInOut{Output}{Output}
		\caption{The Regret Grouping Procedure}
		\label{alg:grouping-t}
		\vspace{0.25\baselineskip}
		
		\Input{$\mathcal{M}$: the set of adjacent 3D segment pairs produced by over-segmentation;\\
		$g$: grouping predictor;\\
		$t$: threshold to stop grouping;\\
		$u$: threshold to enter regret pool;\\
		$k$: number of grouped pairs in each iteration.}
		\Output{The object proposals.}
		\vspace{0.01\linewidth}
		
		Initialize regret pool $\mathcal{R}$ as $\O$; \newline
		
		\While{\rm{True}}{
			Apply $g$ to each 3D segment pair in $\mathcal{M}$;
			
			\If{\rm{The largest} $g$ \rm{output is less than} $t$}{
				Stop grouping and break the loop;
			}
			\Else{
				Select top $k$ pairs with largest $g$ outputs to form a grouping candidate set $\mathcal{M}^{'}$;
				
				\For{\rm{each element} $m^{'}$ \rm{in} $\mathcal{M}^{'}$}{ 
					\If{\rm{$m^{'}$ is in $\mathcal{R}$}}{
						Remove $m^{'}$ from $\mathcal{M}^{'}$; 
					}
					\ElseIf{\rm{regret score of $m^{'}$ is larger than $u$}}{
						Remove $m^{'}$ from $\mathcal{M}^{'}$;
						
						Add $m^{'}$ to $\mathcal{R}$;
					}
				}
				
				Update $\mathcal{M}$ by grouping pairs in $\mathcal{M}^{'}$;
			}
		}
		\Return $\mathcal{M}$;
	\end{algorithm}	
	
	\section{Experiments}
	
	We follow prior work related to 3D objectness \cite{gmu} and take GMU-kitchen dataset as a benchmark for this task, which is more challenging than those widely used datasets of 3D scenes \cite{unsupervised, b3do, scannet}. Furthermore, we find that Cluttered Table-top Dataset \cite{ctd} involves complex object layouts and heavy occlusion. Therefore, it is also taken as a benchmark to make judgment less bias.

	We organize this section as fours parts. Firstly we introduce the two datasets and our evaluation metrics in Sec. \ref{sec:exp-dataset} and Sec. \ref{sec:exp-metrics} respectively. Then we elaborate the details related to our experiments in Sec. \ref{sec:exp-details}. Afterwards, we perform comparative experiments with some state-of-the-art objectness methods from Sec. \ref{sec:exp-gmu} to Sec. \ref{sec:exp-ctd}. Ultimately, we demonstrate the effectiveness of our regret mechanism and grouping predictor in Sec. \ref{sec:abla}).
	
\subsection{Datasets}
	\label{sec:exp-dataset}
	
\begin{figure}[htbp]
	\centering
	\subfigure[GMU-kitchen]{
		\begin{minipage}[t]{0.5\linewidth}
			\centering
			\includegraphics[width=0.99\linewidth]{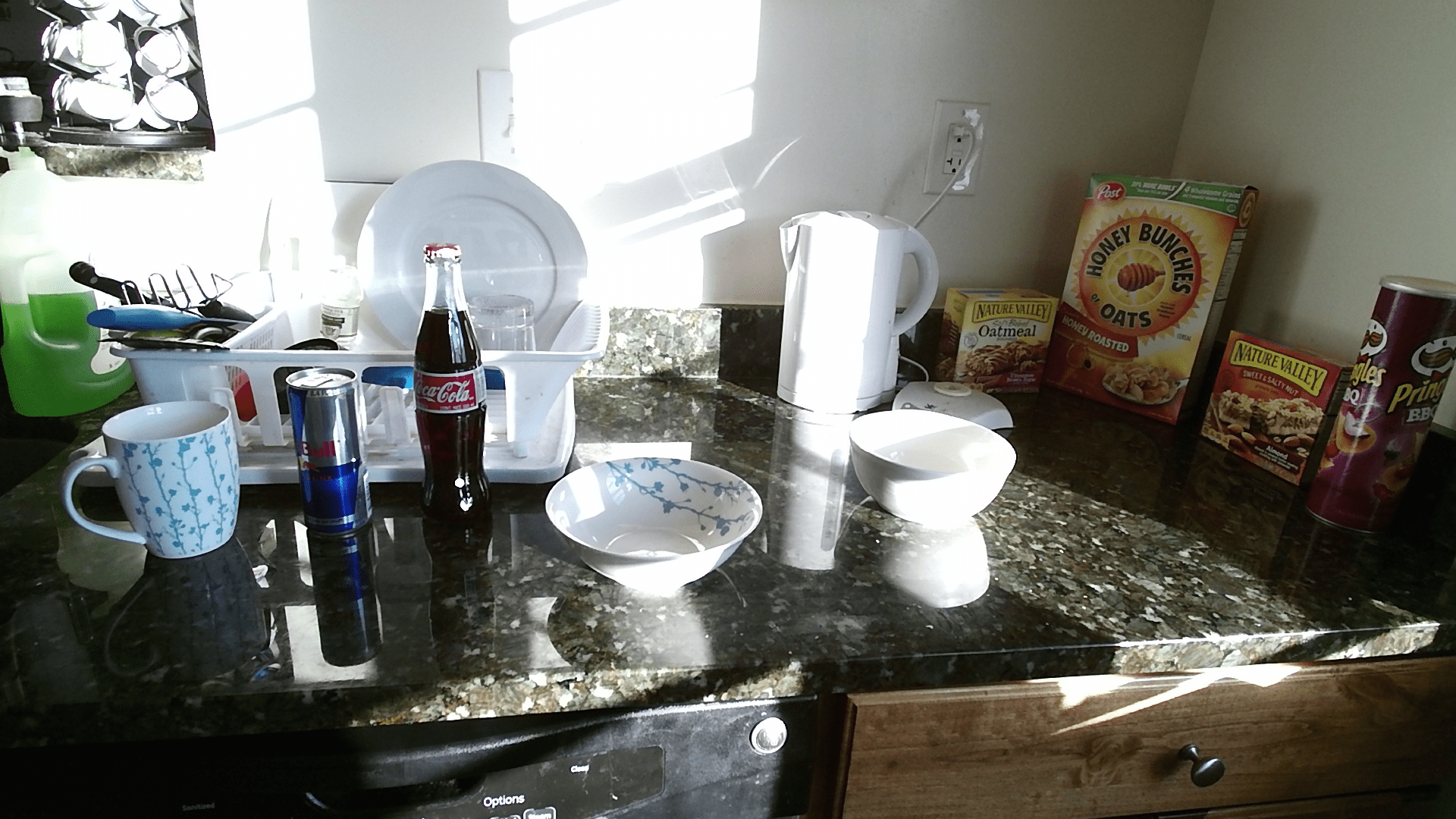}
			
			\vspace{10pt}
			\includegraphics[width=0.99\linewidth]{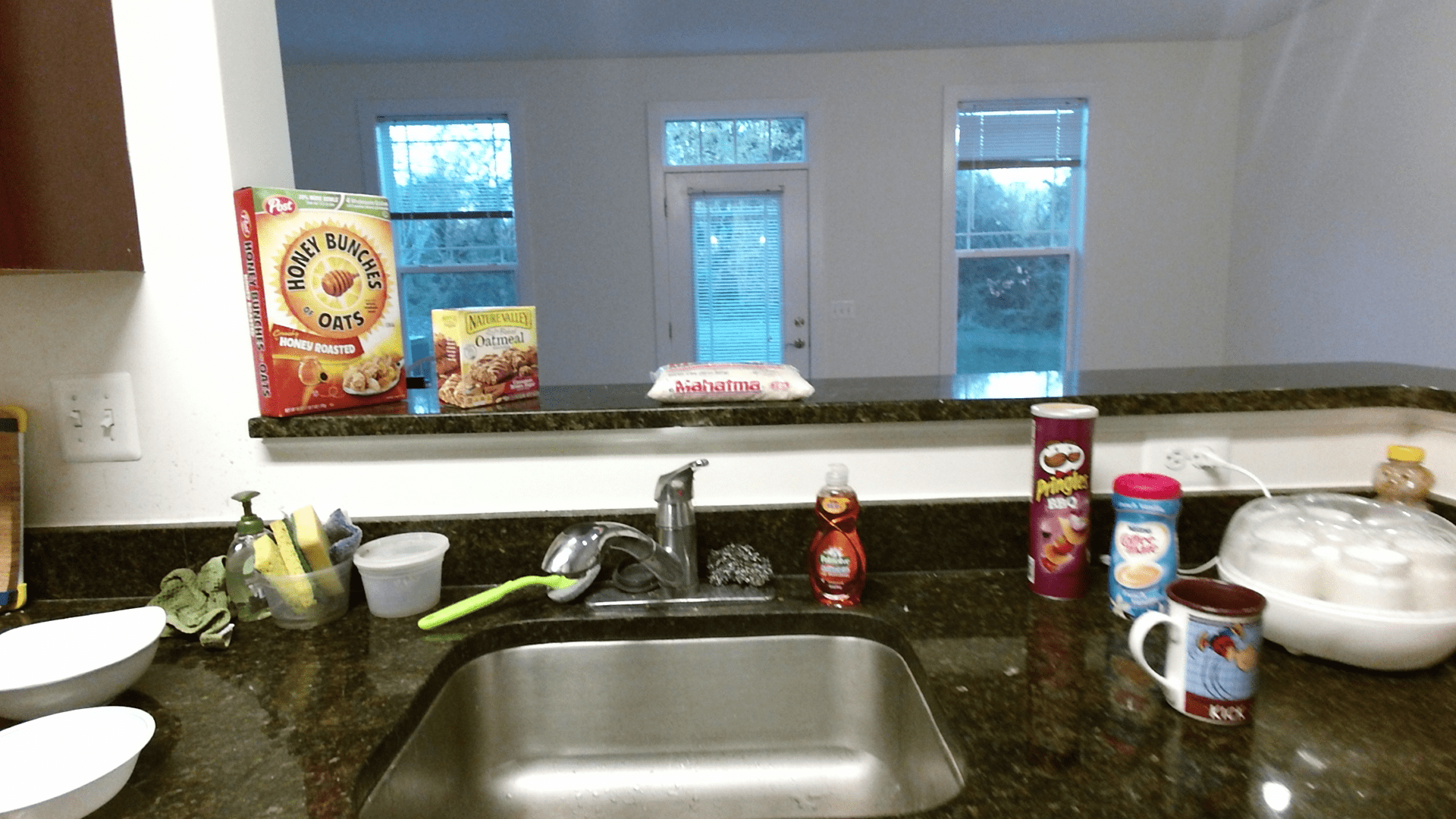}
			
			\vspace{10pt}
			\label{fig:dataset-gmu}
		\end{minipage}
	}
	\subfigure[CTD]{
		\begin{minipage}[t]{0.375\linewidth}
			\centering
			\includegraphics[width=0.99\linewidth]{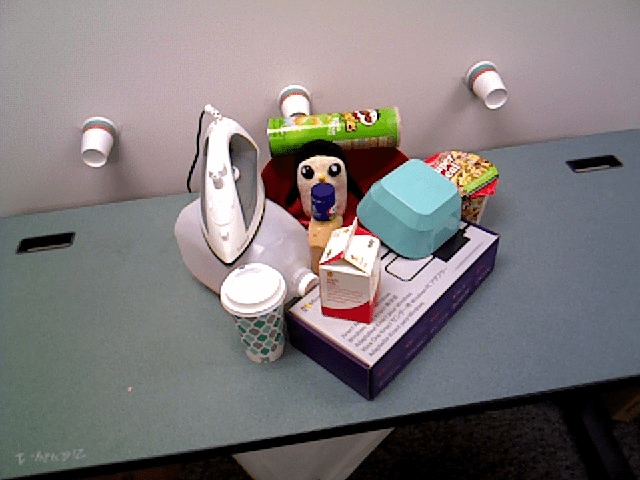}
			
			\vspace{10pt}
			\includegraphics[width=0.99\linewidth]{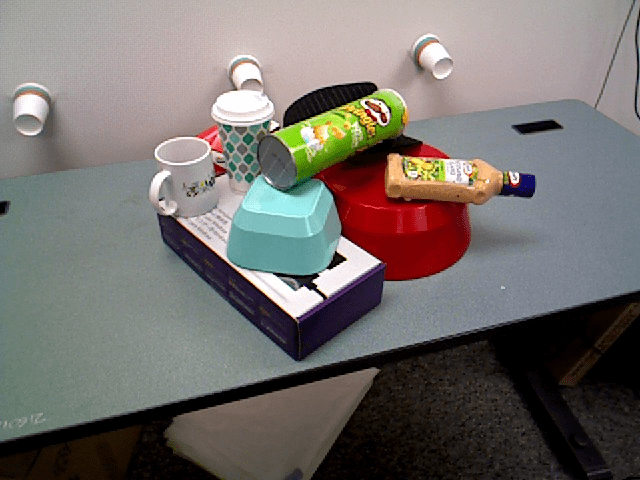}
			
			\vspace{10pt}
			\label{fig:dataset-ctd}
		\end{minipage}
	}
	\centering
	\label{fig:dataset}
	\caption{Typical scenes of the datasets we evaluate on..}
\end{figure}

	\paragraph{GMU-kitchen Dataset.} GMU-kitchen Dataset provides 6735 RGB-D frames and dense point clouds of 9 video sequences from realistic kitchen environments. In each frame, there are more than a dozen common objects in daily life placed in a cluttered fashion. The arrangement of both the background and the objects are relatively complicated (Fig. \ref{fig:dataset-gmu}).
	
	\paragraph{Cluttered Table-top Dataset.} The Cluttered Table-top Dataset (CTD) consists of 3D point clouds and segmentation masks of 89 cluttered scenes with a variety of objects placed on a table. Although each scene contains no more than 10 objects, the high occlusions and complicated geometric structures make the task even more difficult (Fig. \ref{fig:dataset-ctd}).
	
	\subsection{Evaluation Metrics}
	\label{sec:exp-metrics}
	
	We evaluate the performances of objectness methods from the perspective of detection \textit{\textbf{recall}}, namely the radio of objects that have been really detected under a certain IoU overlap threshold.
	
	In general, an object that achieves at least 0.5 IoU with any proposals could be considered detected. To measure the quality of proposals more comprehensively, we also conduct further exploration of the relationship between recall and IoU overlap threshold in our experiments.
	
	Besides, in practice, the number of proposals under a certain recall value should also be considered, because the method which gets certain recall value with fewer proposals can avoid redundancy and greatly improve the efficiency of subsequent processing procedures (\eg robot arm grasping).

	\subsection{Implementation Details}
	\label{sec:exp-details}
	
	\paragraph{Dataset preprocessing.}
	
	Since the GMU-kitchen dataset is generated from video sequences, there exist a large number of redundant frames. To maximize the variance of the point clouds we obtain, we randomly sample 1001 frames with the constraint that overlap between any two frames is no more than 0.8. We construct the point cloud of each scene by projecting depth frame to 3D space according to corresponding camera settings.
	
	For each dataset, we divide the whole data into 70\% training data, 15\% validation data and 15\% test data for training and evaluation. Model selection is conducted on the validation set, and the selected one is then evaluated on the test dataset.
	
	\paragraph{Training scheme.} In general, the scale gaps among 3D segments would be tremendous as the grouping process proceeds, especially when the shape of given objects vary greatly. To this end, we must ensure that the training data covers the 3D segments of all scales. Inspired by \cite{curriculum-learning}, we train our grouping predictor in phases and update the corresponding training set dynamically.
	
	Let $H_i$ denotes the training set of phase-$i$. In phase-1, we build  $H_1$ with the 3D segments obtained from over-segmentation and train the predictor with it, until the average loss drops below a certain threshold $t_{l}$, then proceed to next phase.
	
	Before the training of ($i$+1)-th phase, we perform a selective search iteration, sort all 3D segments pairs according to the prediction results of grouping predictor, then group the top $k$ pairs that belong to the same object according to the ground-truth. Afterwards, we extract new adjacent 3D segments of current state, denoted as $J_{i}$ and construct new training set by $H_{i+1} = H_i \bigcup J_i$. The predictor is then trained with $H_{i+1}$ and step into the next phase when the loss is less than $t_{l}$.
	
	The above procedure is repeated until there is no adjacent 3D segments can be grouped. In this way, we can guarantee that the training set covers all scales of 3D segments, and reduce difficulties compared to training with the data of all phases from scratch.
	
	\paragraph{Parameters settings.}
	
	In our experiments, we optimize grouping predictor using Adam for 1000 epochs with batch size of 32 and initial learning rate of 0.001, and we choose $t_l$ as 0.001.
	
	For regret mechanism, the threshold $u$ for entering regret pool is 0.5, and the number of grouped pairs $k$ in each iteration is 3 for CTD, 10 for GMU-kitchen dataset. In testing, we stop grouping when the largest prediction result is less than 0.75.
	
	\subsection{Evaluation on GMU-kitchen Dataset}
	\label{sec:exp-gmu}
	
\begin{figure}[H]
		\centering
		\subfigure[Recall v.s IoU threshold]{
			\includegraphics[width=0.465\linewidth]{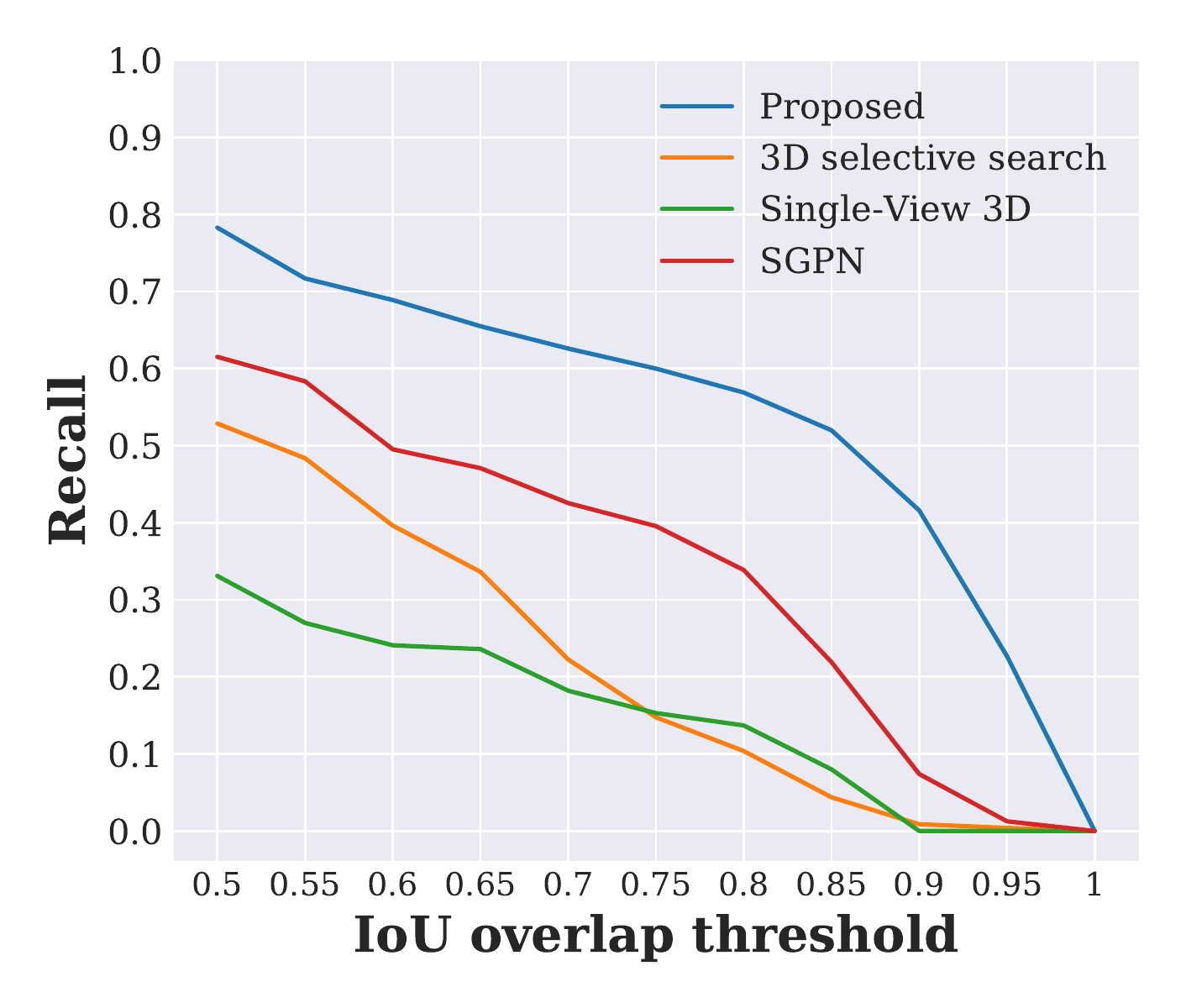}
			\label{fig:gmu-recall-iou-3d}
		}
		\subfigure[Recall v.s number of proposals]{
			\includegraphics[width=0.465\linewidth]{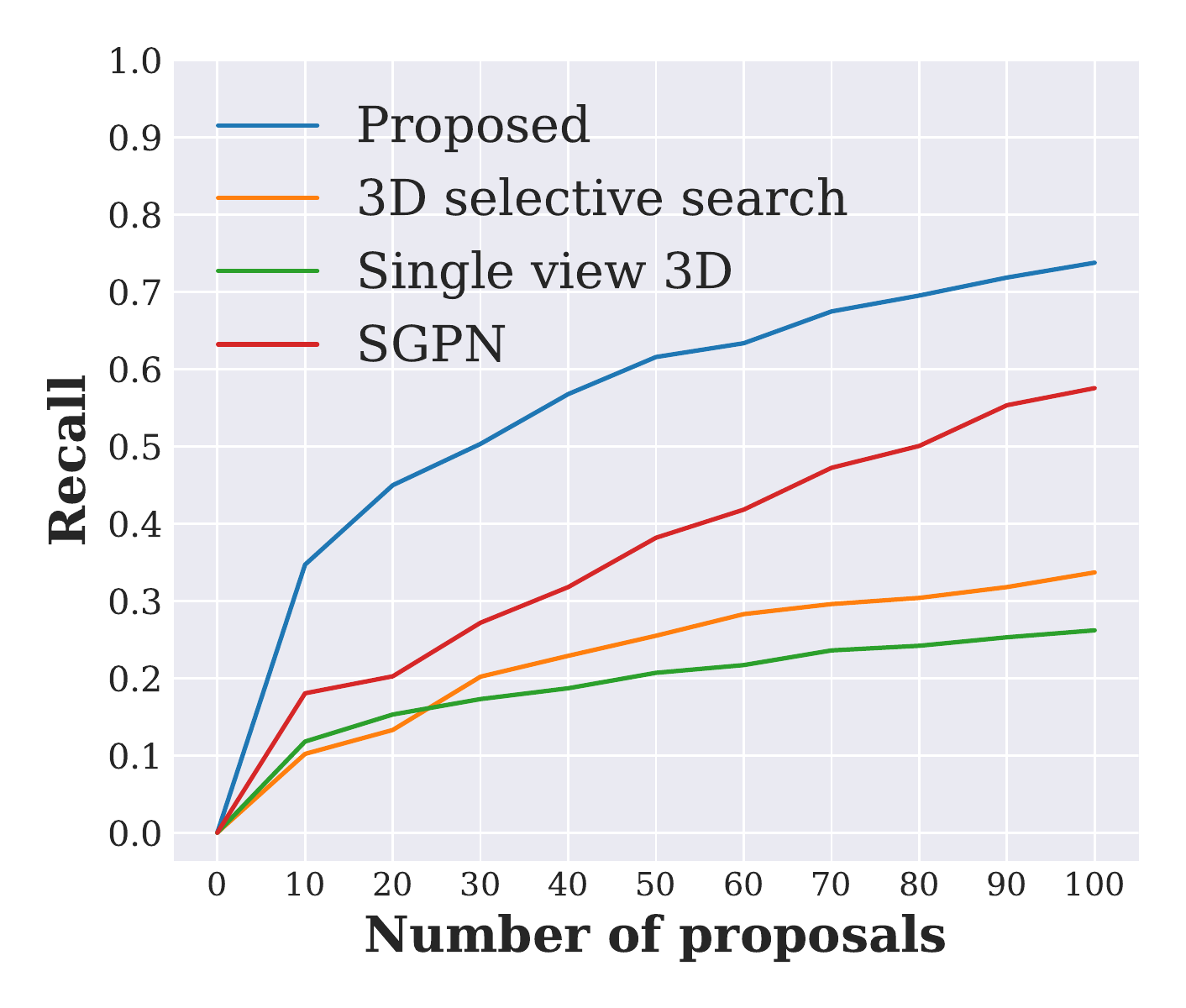}
			\label{fig:gmu-recall-nopro-3d}
		}
		\caption{Evaluation results of 3D objectness methods on GMU-kitchen dataset. We select top 100 proposals to measure the corresponding performances in Fig. \ref{fig:gmu-recall-iou-3d}.}
		\label{fig:gmu-recall-iou}
	\end{figure}

	In the first set of our experiments, we conduct comparisons with three state-of-the-art 3D objectness methods, 3D Selective Search (3D-SS) \cite{3d-ss}, Single-View 3D \cite{gmu} and SGPN \cite{sgpn} on GMU-kitchen dataset. Note that the dense point clouds are usually difficult to obtain in practice, thus we only consider Single-View 3D instead of Multi-View 3D in \cite{gmu}. The results are illustrated in Fig. \ref{fig:gmu-recall-iou}.
	
	We observe that our method outperforms the other three 3D objectness methods by a considerable margin (See Fig. \ref{fig:gmu-recall-iou-3d}), especially at high IoU threshold. The performances of both 3D-SS and Single-View 3D degrade significantly as the IoU threshold increases.
	
	On the contrary, our method tends to maintain a relatively high recall under 0.5-0.9 IoU thresholds. This suggests that our objectness method generates high quality proposals that fit the ground-truth well while most proposals of the other two 3D objectness methods have poor overlap with the objects.
	
	Based on the embedding of point-wise contextual features, SGPN seems to learn objectness well and shares the similar trend with our method in Fig. \ref{fig:gmu-recall-iou-3d}. Nevertheless, due to the lack of regret mechanism, its performance is still limited by mistaken grouping operations in early merging phases.
	
	Fig. \ref{fig:gmu-recall-nopro-3d} indicates that our method is capable of performing on par with the other three methods under less proposals. The traditional objectness methods tend to achieve ideal performance at the expense of a large number of proposals (especially for 3D-SS).

	\subsection{Evaluation on CTD}
	\label{sec:exp-ctd}
	
\begin{figure}[htbp]
		\centering
		\subfigure[Recall v.s IoU threshold]{
			\includegraphics[width=0.465\linewidth]{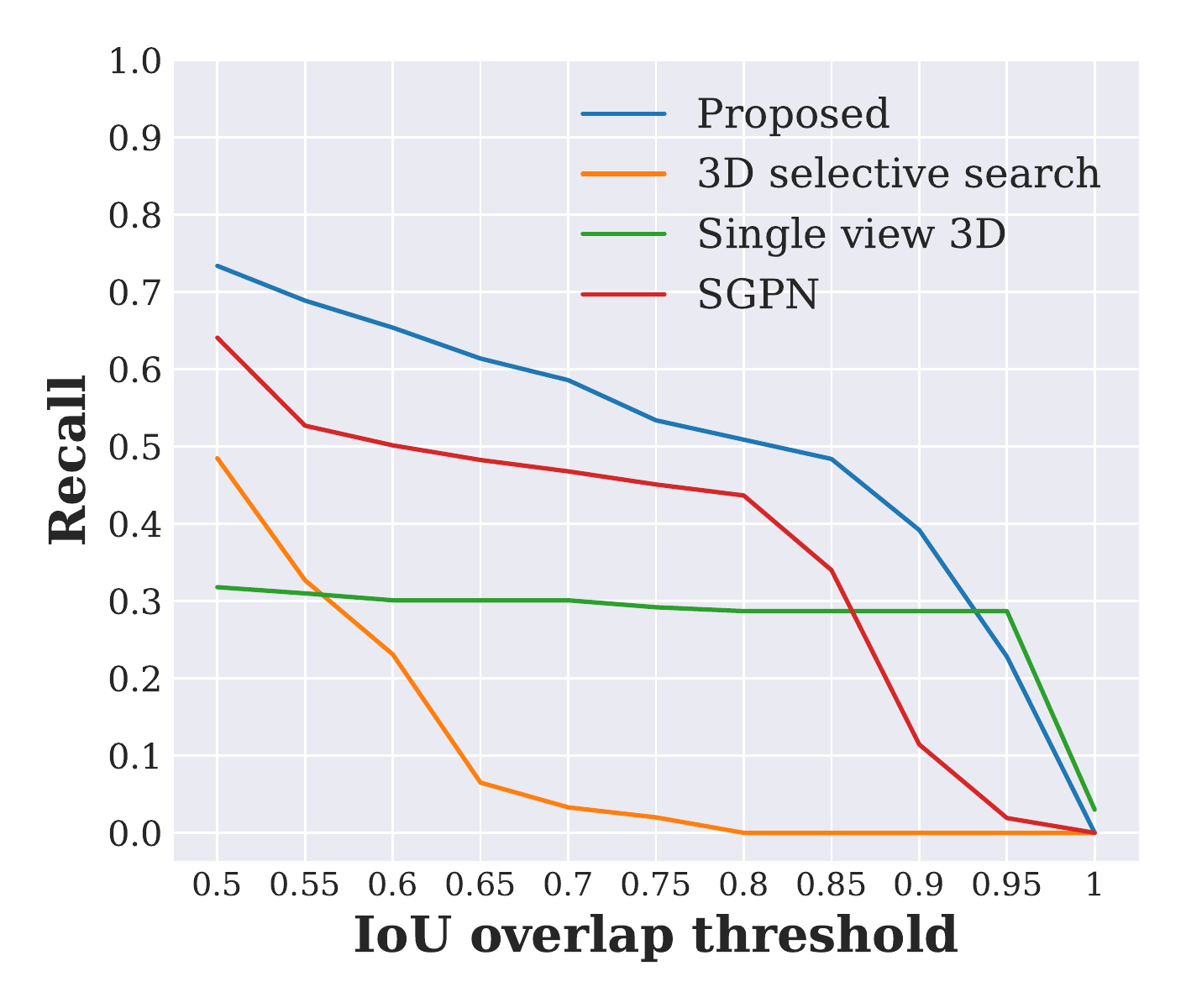}
			\label{fig:ctd-recall-iou-3d}
		}
		\subfigure[Recall v.s number of proposals]{
			\includegraphics[width=0.465\linewidth]{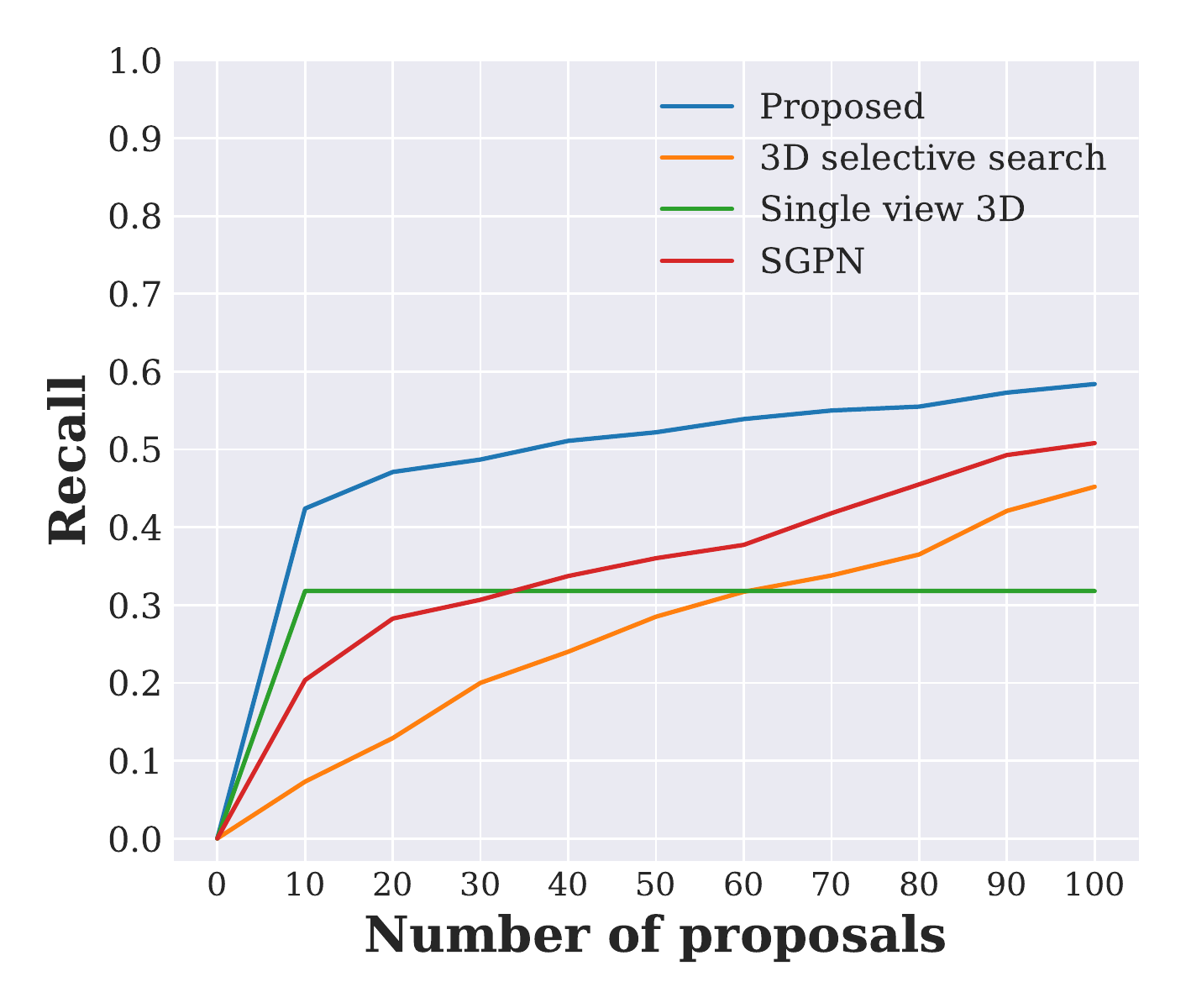}
			\label{fig:ctd-recall-nopro-3d}
		}
		\caption{Evaluation results of 3D objectness methods on CTD. We select top 200 proposals to measure the corresponding performances in Fig. \ref{fig:ctd-recall-iou-3d}.}
		\label{fig:ctd-recall-iou}
	\end{figure}

	To further explore robustness of our method under clutter scenes and complicated geometric structures, we conduct the same evaluation on CTD as in Sec. \ref{sec:exp-gmu} and Fig. \ref{fig:ctd-recall-iou} shows the comparative results.

    Actually, objectness on CTD is much more challenging than GMU-kitchen dataset due to the high occlusion and complex geometric shapes. As illustrated in Fig. \ref{fig:ctd-recall-iou}, our method also shows its robustness and outperforms the compared methods in all regimes.
    
    For such occluded scenes, our method and SGPN can extract the contextual features between object instances better and achieve satisfying performances with deep learning schemes. On the contrary, traditional 3D objectness methods would encounter difficulties, the performance of 3D-SS declines rapidly as the IoU overlap threshold increases and Single-View 3D tends to crash for always mistaking the cluttered scene to an object.

\begin{figure*}[htbp]
	\centering
	\subfigure[0\%]{
		\begin{minipage}[t]{0.185\linewidth}
			\centering
			\includegraphics[width=1.1in]{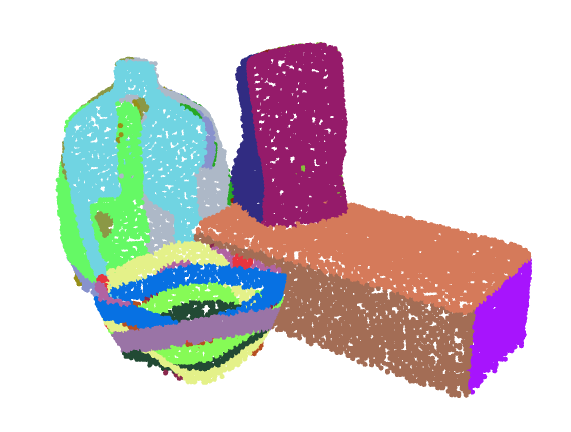}
			
			\vspace{10pt}
			\includegraphics[width=1.1in]{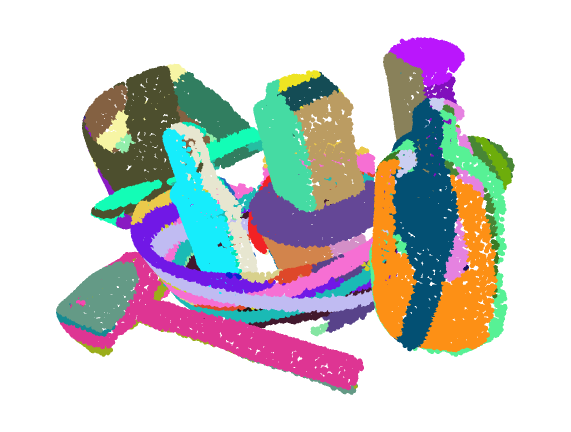}
			
			\vspace{10pt}
			\includegraphics[width=1.1in]{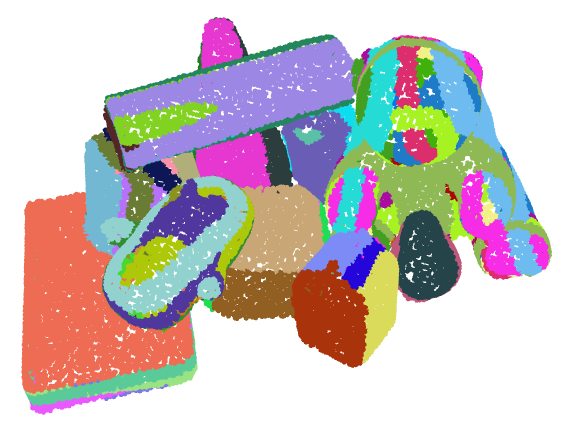}
			
			\vspace{12pt}
		\end{minipage}
	}
	\subfigure[25\%]{
		\begin{minipage}[t]{0.185\linewidth}
			\centering
			\includegraphics[width=1.1in]{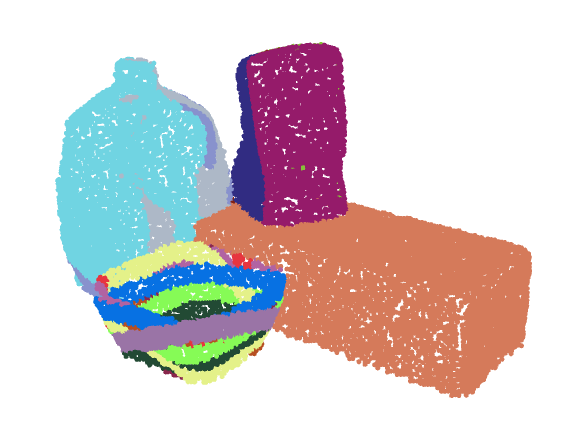}
			
			\vspace{10pt}
			\includegraphics[width=1.1in]{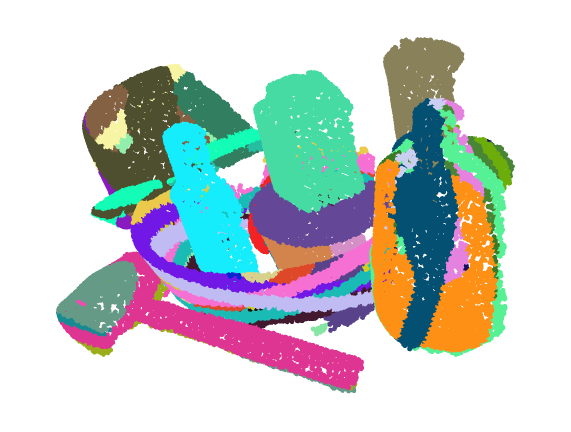}
			
			\vspace{10pt}
			\includegraphics[width=1.1in]{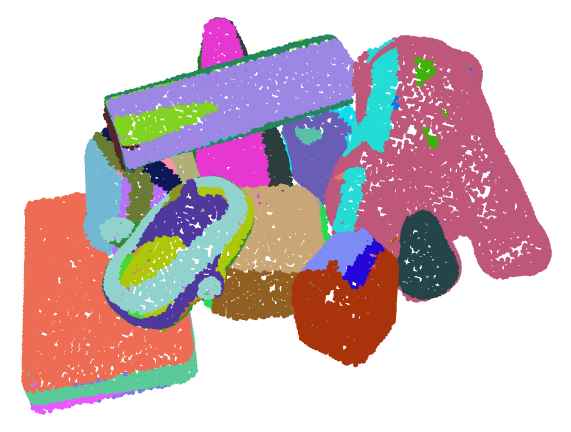}
			
			\vspace{12pt}
		\end{minipage}
	}
	\subfigure[50\%]{
		\begin{minipage}[t]{0.185\linewidth}
			\centering
			\includegraphics[width=1.1in]{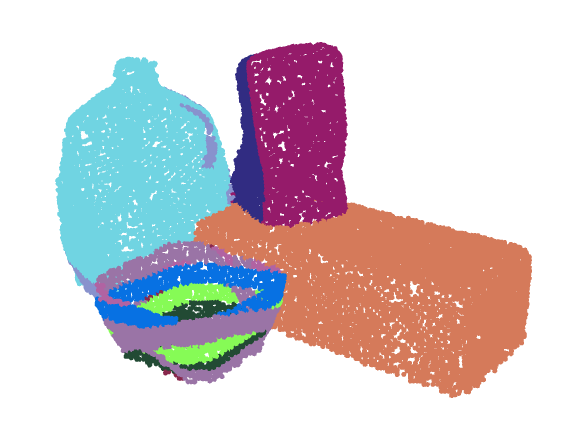}
			
			\vspace{10pt}
			\includegraphics[width=1.1in]{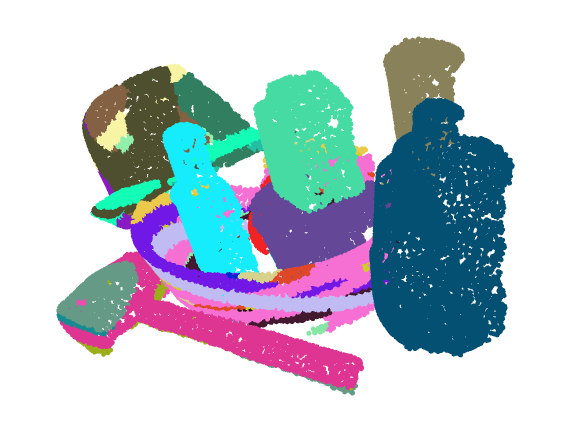}
			
			\vspace{10pt}
			\includegraphics[width=1.1in]{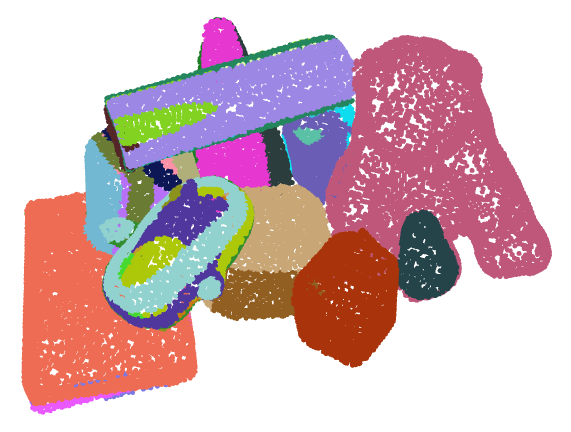}
			
			\vspace{12pt}
		\end{minipage}
	}
	\subfigure[75\%]{
		\begin{minipage}[t]{0.185\linewidth}
			\centering
			\includegraphics[width=1.1in]{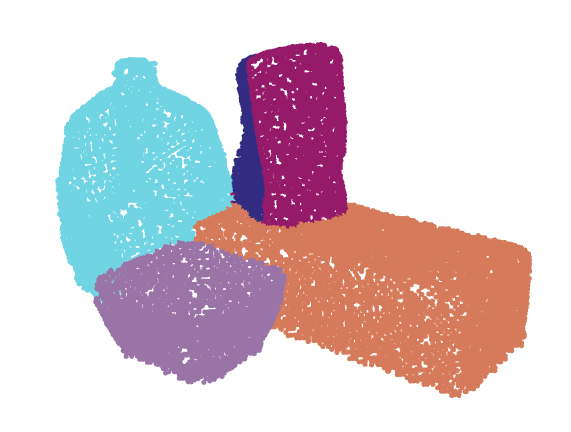}
			
			\vspace{10pt}
			\includegraphics[width=1.1in]{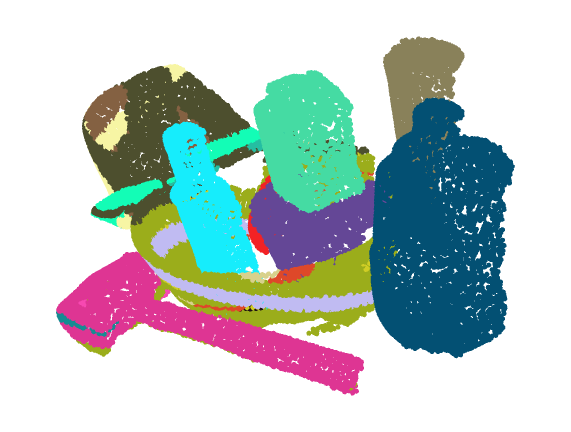}
			
			\vspace{10pt}
			\includegraphics[width=1.1in]{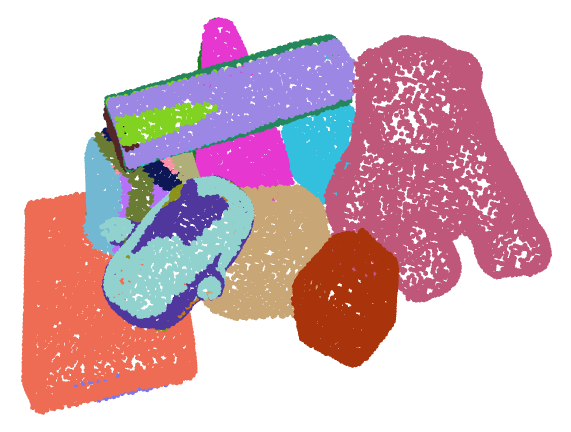}
			
			\vspace{12pt}
		\end{minipage}
	}
	\subfigure[100\%]{
		\begin{minipage}[t]{0.185\linewidth}
			\centering
			\includegraphics[width=1.1in]{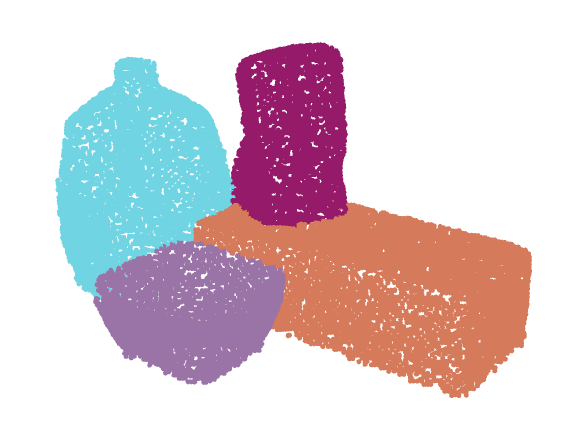}
			
			\vspace{10pt}
			\includegraphics[width=1.1in]{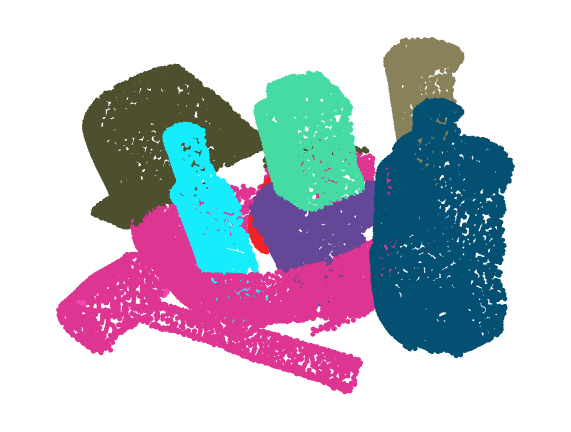}
			
			\vspace{10pt}
			\includegraphics[width=1.1in]{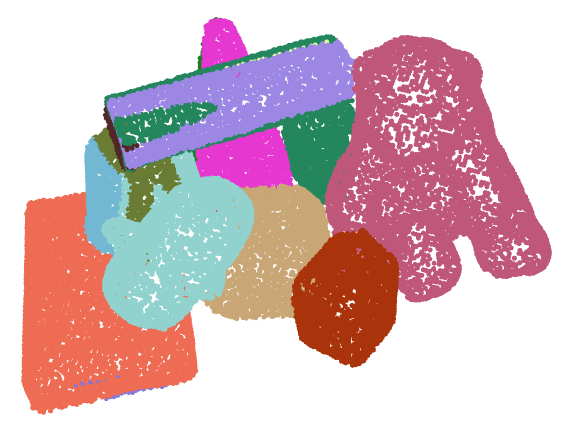}
			
			\vspace{12pt}
		\end{minipage}
	}
	\centering
	\label{fig:supp-exp}
	\caption{The grouping process of some typical scenes on CTD. Each row denotes a scene and the numbers below each column represent the corresponding grouping schedules. The leftmost column illustrates the over-segmentation results and the right most column shows the final object proposals.}
\end{figure*}

    To present the grouping process of our objectness method more clearly, we further visualize it with some typical scenes in Fig. \ref{fig:supp-exp}.
	
	\subsection{Ablation Study}
	\label{sec:abla}
	
	In Sec. \ref{sec:regret}, we propose a regret grouping mechanism. Due to the high occlusion of scenes and complicated object shapes in CTD, incorrect grouping operations are more likely to occur, where regret mechanism might show its strength. On the other hand, we also introduce a grouping predictor to substitute the traditional hand-crafted similarity metrics. Therefore, to ascertain the effectiveness of the two components, we conduct extensive ablation experiments on CTD.
	
	We analyze their effects under three conditions: 1) using regret mechanism and grouping predictor; 2) using predictor without regret mechanism; 3) using neither regret mechanism nor predictor (\eg 3D selective search).
	
	\begin{figure}[htbp]
		\centering
		\includegraphics[width=0.8\linewidth]{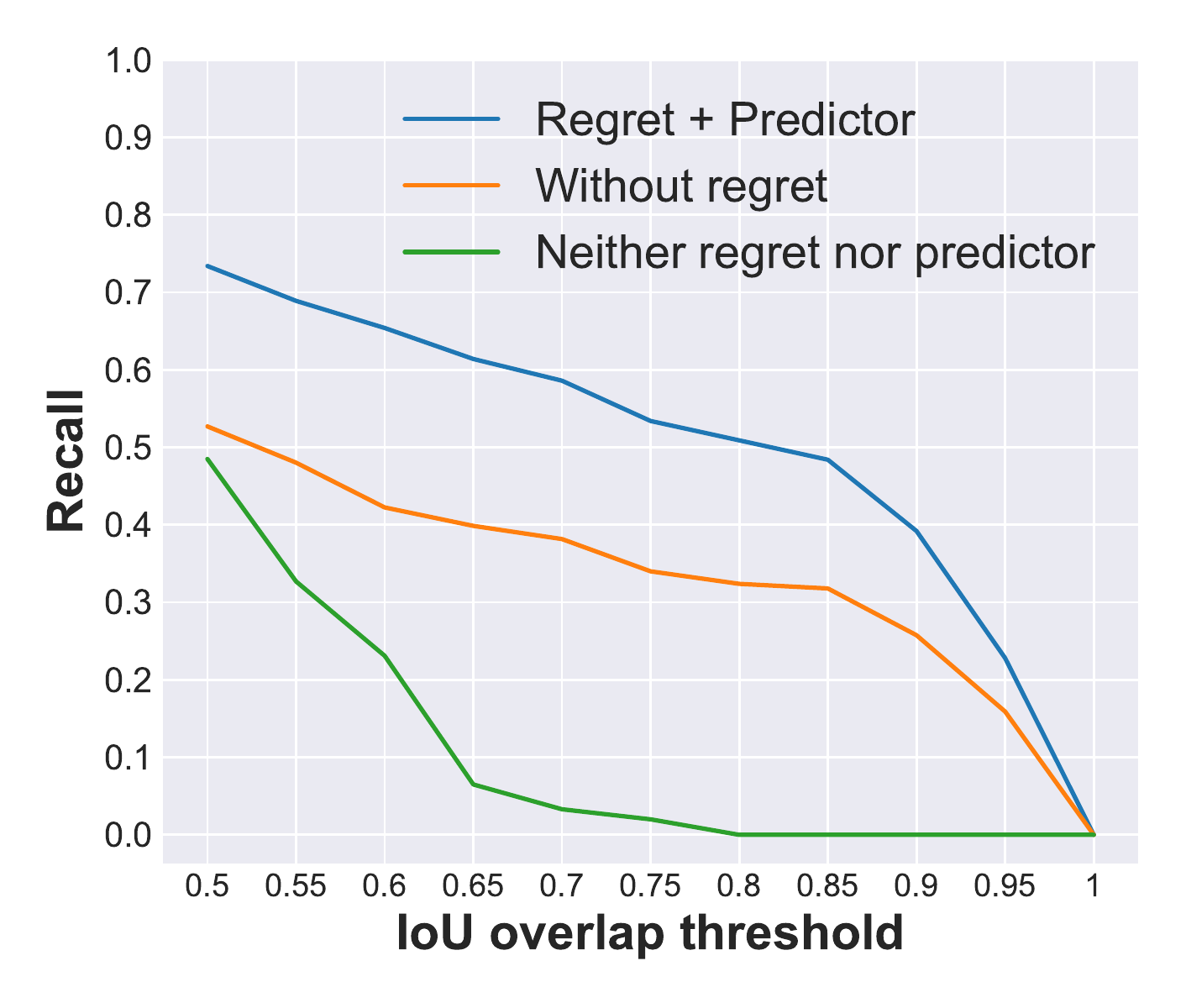}
		\caption{Recall vs IoU threshold curves of ablation experiments on CTD.}
		\label{fig:ctd-abla}
	\end{figure}
	
	As is illustrated in Fig. \ref{fig:ctd-abla}, both regret mechanism and the predictor play significant roles. Regret mechanism can overwhelmingly improve the overall performance, while grouping predictor would modify the trend between recall and IoU overlap threshold and lead to better performance under high IoU overlap threshold.
    
    \paragraph{Transferability.} Our grouping predictor is learned from basic geometric structures instead of the whole object instances. Most objects from different datasets share similar over-segmentation results, thus the learned model can be transferred between them. We also train our model on GMU-kitchen dataset and test it in CTD (unseen objects for GMU-kitchen dataset). We achieve \textbf{63.2\%} recall under 0.5 IoU overlap threshold, which is still superior to 3D-SS and Single-view 3D (48.5\% and 31.8\%) and on par with SGPN (64.1\%). Note that the compared methods are both trained and tested on CTD.
    
    \paragraph{Computation time.} In our experiments, our speed performance is decent. We have recorded the average execution time on the test set of CTD. The program is run on the Xeon E5-2680 and a GTX 1080Ti, taking about 400ms per scene. Actually, the number of grouping iterations is not that large and PointNet is one of the most simple and efficient networks to process 3D point cloud data.
	
	\section{Discussion and Future Work}
	Our predictor estimates the grouping probability only based on the feature vectors of current 3D segment pairs. For two 3D segments pairs with the same spatial structure, our predictor tends to output the same result, whereas the corresponding ground-truth might be different, thus leading to ambiguity.
	
	The occurrence of such case would be affected by the distribution of training data and can be largely circumvented via regret mechanism. However, we argue that this could be further improved via taking advantage of previous grouping results. It would be interesting to explore some modification on the architecture of predictor (\eg RNN, GCN) to address this issue. It may also be a good option to introduce different training schemes (\eg reinforcement learning).
	
	\section{Conclusion}
	
	We have presented a bottom-up objectness method on raw 3D point clouds. Given 3D point cloud of the scene, an over-segmentation was conducted to produce a set of 3D segments. Then a grouping predictor learned from a large amount 3D object models was utilized to determine whether a pair of 3D segments should be grouped, which avoided the labor of exploring similarity metrics and promoted performance greatly. By introducing a novel regret mechanism, we largely addressed the issue of conventional bottom-up methods due to greedy grouping. The experiments on GMU-kitchen dataset and CTD demonstrated that our 3D objectness method outperformed state-of-the-art 3D objectness methods with remarkable margins and was significantly more robust against extreme cases (\eg occlusion, complex object models) with the aid of our regret mechanism.

{\small
\bibliographystyle{ieee_fullname}
\bibliography{egbib}
}
\clearpage

\end{document}